\documentclass{article}

\usepackage{amsmath}
\usepackage{graphicx}
\usepackage{subcaption}
\usepackage{makecell}
\usepackage{wrapfig}

\usepackage[preprint]{neurips_2026}

\usepackage[utf8]{inputenc} 
\usepackage[T1]{fontenc}    
\usepackage{hyperref}       
\usepackage{url}            
\usepackage{booktabs}       
\usepackage{amsfonts}       
\usepackage{nicefrac}       
\usepackage{microtype}      
\usepackage{xcolor}         

\title{Unsupervised Latent Space Alignment with Hyperspherical Geodesic Matching}

\author{%
  Cameron Ryan\thanks{Northeastern University, Boston, MA 02115.
    Work performed during a summer internship at Lawrence Livermore National Laboratory} \\
  \texttt{ryan.came@northeastern.edu} \\
  \And
  Vivek Sivaraman Narayanaswamy\thanks{Lawrence Livermore National
    Laboratory, Livermore, CA 94551-0808.} \\
  \texttt{narayanaswam1@llnl.gov} \\
  \And
  Kowshik Thopalli\footnotemark[2] \\
  \texttt{thopalli1@llnl.gov} \\
  \And
  Shusen Liu\footnotemark[2] \\
  \texttt{liu42@llnl.gov} \\
}

\begin{document}

\maketitle

\begin{abstract}
  Independently trained neural networks tend to encode the same data with similar latent geometries. 
These latent geometries are not directly compatible, yet they can be nearly the same up to some class of transformations.
While there exists many methods for alignment between different latent spaces, it is typically done using a set of shared sample correspondences, known as \emph{anchors}.
This leaves a fundamental question: are the geometric signatures of different latent spaces representing similar data sufficient to recover an alignment between them?
To that end, we introduce HGA (\emph{Hyperspherical Gaussian Alignment}), a method that directly optimizes a transformation between two latent spaces by maximizing a geometric measure of ``fit'' between them. 
Since it is driven by the geometry of the latent spaces rather than paired data, HGA can operate in both an unsupervised and weakly supervised regime. 
On tasks such as model stitching or multilingual word embedding correspondence recovery, HGA manages to match supervised results with minimal or no supervision. 
\end{abstract}
\section{Introduction}

Artificial neural networks often condense extremely high dimensional data points into much lower dimensional latent spaces.
The structure and geometry of datasets embedded in these latent spaces is central to how models are used, compared, or reused. 
A principle for why latent spaces are able to represent data is the \emph{manifold hypothesis}, which states that most natural high dimensional data lies on a manifold with a much lower intrinsic dimension \cite{fefferman2013testingmanifoldhypothesis, bengio2014representationlearningreviewnew}.
In this view, a trained model implicitly captures the geometry of the data manifold in its latent space.

Evidence suggests that models trained independently on similar data often learn the same representations up to certain transformations such as isometries and rescaling \cite{moschella2023relative, ainsworth2023git, theus2025generalized}.
The Platonic Representation Hypothesis \cite{pmlr-v235-huh24a} posits that, as models scale and train on larger datasets, their representations converge toward a statistical structure of the underlying world. 
Moreover, much work has investigated translation between latent spaces, often using supervision in the form of sample pairs called \emph{anchors}.
These act like a ``Rosetta Stone'', providing known correspondences between two latent spaces \cite{moschella2023relative, fumero2024latent}. 
It has been clearly demonstrated that using anchors, effective latent space alignment is possible, and that often latent spaces are the same up to an affine or orthogonal transformation. 
The central question that motivates this work is \emph{whether compatible latent spaces can be aligned by relying on their geometric signatures alone?}
Answering this question requires methods to align spaces by exploiting their intrinsic geometric structure rather than using paired data.

We address this challenge with Hyperspherical Geodesic Alignment (HGA), which attempts to align latent spaces by fitting an orthogonal transformation that maximizes similarity between the mapped space and the target space. 
HGA is motivated by the assumption that in separate latent spaces representing similar data, the relative angles between corresponding points are approximately the same \cite{moschella2023relative}. 
We find that in many cases, supervision is not necessary and the intrinsic geometry of the latent spaces alone is enough to determine an alignment between them.
Our key contributions are summarized as follows:
\begin{itemize}
    \item Propose HGA, a method for latent space alignment in both the unsupervised and weakly supervised regimes, demonstrating that in many cases, their intrinsic geometry is sufficient to align them.

    \item Empirically compare HGA to Latent Functional Maps \cite{fumero2024latent} and fully supervised Orthogonal Procrustes fitting across several datasets and data modalities, showing that it improves alignment quality with no or significantly less supervision.
\end{itemize}
\section{Related Work}
\noindent \textbf{Representational similarity and convergence.}
Prior work shows that the internal representations of neural networks trained under different conditions with similar data will tend to be similar, and nearly equivalent up to certain transformations \cite{raghu2017svccasingularvectorcanonical, pmlr-v97-kornblith19a}.
Often, a useful way to quantify the compatibility of different representations is by \emph{model stitching}.
Model stitching involves connecting lower layers of one neural network with the upper layers of another neural network with a learned transformation, and quantifies the stitched network's performance at a downstream task \cite{lenc2015understandingimagerepresentationsmeasuring, bansal2021revisitingmodelstitchingcompare}.
On a higher level, the Platonic Representation Hypothesis \cite{pmlr-v235-huh24a} contends that as models grow and are trained on more data and broader tasks, their representations converge toward a single statistical structure of the physical world. 
The observations of these prior works motivate our setting: \emph{do latent spaces have geometric signatures that can facilitate alignment between similar latent representations?} 
Our work differs from the previous line of research because rather than attempting to \emph{measure} similarity, we attempt to \emph{exploit} it for alignment.
    
\noindent \textbf{Weak-supervision latent space communication.}
There exists a line of work that seeks to translate between latent spaces using minimal supervision, typically in the form of a set of shared samples called \emph{anchors}.
Relative Representations \cite{moschella2023relative} propose representing points in each latent space as their similarities to a fixed set of anchors.
Therefore this method is invariant to latent isometries that otherwise separate independently trained models, and thus enables zero-shot communication between latent spaces.
Subsequent work \cite{maiorca2024latentspacetranslationsemantic} brings this further by directly estimating a transformation between the two spaces so that encoders and decoders can be combined.
Latent Functional Maps \cite{fumero2024latent} introduces a method for latent space communication based on spectral geometry, notably the \emph{Functional Maps} framework \cite{ovsjanikov2012functional}.
This method aligns graph Laplacian eigenvectors to preserve the spectral features of indicator functions of anchor points.
Latent Functional Maps also introduces a method of unsupervised latent space communication, by optimizing a map between the spectral spaces of two latent space such that it preserves the Heat Kernel Signature. 
Our approach is within this weakly supervised or unsupervised regime.
We optimize a transformation between two latent spaces that will maximize a geometric ``fit" between the mapped space and the target space. 

\noindent \textbf{Optimal Transport.}
Optimal transport (OT) compares probability measures by discovering a minimum cost coupling between them \cite{cuturi2013sinkhorndistanceslightspeedcomputation, peyre2019computational}, and has been used to compare and align embedding spaces \cite{pmlr-v37-kusnerb15}.
Classic OT requires a cost defined between two embedding spaces which is unavailable when embeddings are related up to an unknown transformation.
Gromov Wasserstein (GW) distance \cite{memoli2011} overcomes this by instead comparing \emph{intra}-space pairwise distances. 
This yields a coupling which preserves the relational structure between points. 
Prior work has used GW to find alignments between word embedding spaces for different languages without anchors \cite{alvarez_2018, zhang-etal-2017-earth}.
These approaches are highly related to this work, as both seek to find an optimal way to align high dimensional spaces by preserving their internal geometry. 
Our work differs slightly because rather than a point-wise coupling, we optimize an orthogonal transformation (therefore automatically preserving intra-space structure) which maximizes the alignment of dense regions to other dense regions.
\newcommand{\M}{\mathcal{M}}
\newcommand{\N}{\mathcal{N}}
\newcommand{\din}{\rotatebox[origin=c]{270}{\in}}

\section{Method}
\noindent \textbf{Setup.}
Consider two sets of points sampled from two latent spaces: $S_\M\subset \M\subset\mathbb{R}^{m\times d}$, and $S_\N\subset\N\subset\mathbb{R}^{n\times d}$.
An important result of \cite{moschella2023relative} is demonstrating that different neural networks will typically represent data points so that their angles between each other are nearly the same.
Our method seeks to take advantage of this by deliberately rotating/reflecting these samples in order to find the best possible ``fit'' between them.

\noindent\textbf{Hyperspherical Geodesic Alignment (HGA).} Consider these datasets normalized to the unit sphere $\hat{S}_\M$ and $\hat{S}_\N$. Let $R\in O(d)$ be an orthonormal matrix. Our goal is to optimize $R$ such that we maximize the ``fit'' between $R\hat{S}_\M$ and $\hat{S}_\N$. We calculate the cosine similarity between each point in $S_\M$ after being operated on by $R$, and each point in $S_\N$:
\[
\underset{\substack{\rotatebox[origin=c]{270}{$\in$} \\ \mathbb{R}^{m\times d}}}{(R\hat{S}_{\mathcal{M}}^{T})^{T}}
\underset{\substack{\rotatebox[origin=c]{270}{$\in$} \\ \mathbb{R}^{d\times n}}}{\hat{S}_{\mathcal{N}}^{T}}
\in \mathbb{R}^{m\times n}
\]
We can then calculate the geodesic distance between each point in $\M$ and each point in $\N$: 
    \[
    \arccos((R\hat{S}_\M^T)^T\hat{S}_\N^T)\in \mathbb{R}^{m\times n}
    \]
Using geodesic distances on the unit sphere, we calculate the unnormalized Gaussian Kernel between each point:
    \[
    \exp
    \left(
    -\frac{
    \arccos^{2}
    \big(\!\big(R\hat{S}_{\mathcal{M}}^{T}\big)^{T}\hat{S}_{\mathcal{N}}^{T} \big)_{ij}
    }
    {2\sigma^2}
    \right)
    \]
We then average the sum over all pairs of points between the two spaces. This total affinity is our objective function:
\[
    J_\sigma(R)=
    \frac{1}{mn}
    \sum_{i,j}
    \exp
    \left(
    -\frac{
    \arccos^{2}
    \big(
    \!\big(
    R\hat{S}_{\mathcal{M}}^{T}
    \big)^{T}\hat{S}_{\mathcal{N}}^{T} 
    \big)_{ij}}{2\sigma^2}\right)
\]
The goal of our optimization algorithm is find $\text{argmax}_{R\in O(d)}J_\sigma(R)$.
Intuitively, we imagine this as slowly rotating the points of $\hat{S}_\M$ until they ``fit'' together with the points of $\hat{S}_\N$.
We find an ideal $R$ matrix by starting from a random orthogonal matrix, and progressively updating it with gradient ascent.
Details are elaborated in the Appendix.

\noindent\textbf{Supervision.} Methods for latent space alignment typically use a set of anchor points $A=\{x_{a_1},\dots, x_{a_N}\}\subset\mathcal{D}$ and optimize a transformation $T:\M\to \N$ to satisfy $T(S_\M(x_{a_i}))\approx S_\N(x_{a_i})$.
These points are often considered to be a ``Rosetta Stone'' between latent spaces \cite{maiorca2024latentspacetranslationsemantic, norelli2023asifcoupleddataturns}.
In our experiments, we investigated two methods for implementing anchor supervision in HGA.
These two methods are often combined. 

\underline{\textit{Starting From a Least Squares Fit.}} The orthogonal Procrustes problem has been well understood and studied \cite{Schonemann_1966}. Since our optimization algorithm requires an initialization of $R$, we find the best fit at the anchor points: $ \min_{R} \| \hat{S}_\M(A) R - \hat{S}_\N(A) \|$.
We use this estimated $R$ as the initialization for the optimization algorithm.

\underline{\textit{Anchor Weighting.}} The anchor relationships can also be enforced by giving them higher weights in the objective function sum. 
If we know that $(\hat{S}_{\M})_i$ and $(\hat{S}_{\N})_j$ correspond to the same anchor in the two respective latent spaces, then we can weight their relationship higher. 
Thus, the objective function becomes:
\[
J_\sigma(R)=
\sum_{i,j}\alpha_{ij}\exp\!
\left(
-\frac{\arccos^{2}\!
\big(
\big(
R\hat{S}_{\mathcal{M}}^{T}
\big)
^{T}\hat{S}_{\mathcal{N}}^{T}
\big)_{ij}}{2\sigma^{2}}
\right),\qquad \alpha_{ij}=\begin{cases}w, & (i,j)\text{ is anchored},\\1, & \text{otherwise}.\end{cases}
\]

An important consideration in experiments is the choice of the weight hyperparameter $w$. 
A value of $w$ that is too small will not effectively change the objective landscape, while a value that is too large can make it unstable. 
Our experiments show that it is often helpful to set $w\approx m\times n$ so that it is larger than the total number of connections.

\section{Experiments}

\noindent\textbf{Model Stitching.}
A common challenge in latent communication tasks is combining components of two different neural networks to create a stitched model \cite{lenc2015understandingimagerepresentationsmeasuring, bansal2021revisitingmodelstitchingcompare, csiszarik2021similarity}. Consider neural networks $f(x)=f_{L_f}\circ f_{L-1}\circ\cdots\circ f_2\circ f_1 (x)$, and $g(x)=g_{L_g}\circ g_{L-1}\circ \cdots \circ g_{2}\circ g_1(x)$.
In order to stitch layer $i$  in $f$ to layer $j$ in $g$, we search for a transformation $T:\mathbb{R}^{d_1}\to \mathbb{R}^{d_2}$ so that $g_{L_g}\circ \cdots \circ g_{j+1}\circ T\circ f_{i}\circ \cdots \circ f_1(x)$ is sufficiently good at the downstream task which $g$ is used for.
Previous work \cite{maiorca2024latentspacetranslationsemantic, moschella2023relative, moayeri2023texttoconceptandbackcrossmodel} has shown that often orthogonal or affine transformations are sufficient to map between the latent spaces of the corresponding layers in a neural network. For simplicity, in this work we focus on orthogonal transformations between the latent spaces of the corresponding layers. 

Stitching methods often use a set of anchor points $\{x_{a_1},\dots, x_{a_N}\}\subset\mathcal{D}$, which are used to direct the transformation to preserve similarities between how the points are represented, so that $f_i\circ\cdots\circ f_1(x_{a_k})\approx g_j\circ\cdots\circ g_1(x_{a_k})$. 
Our model stitching experiments focus primarily on image classification for the  \cite{mnist}, CIFAR-10, and CIFAR-100 \cite{krizhevsky2009learning} datasets.
We train pairs of identical models on each dataset, and perform the optimization procedure using a subset of 1024 samples to find an alignment between their respective latent spaces.
We find that in many cases, the usage of anchor supervision is unnecessary for model stitching.
By observing Table \ref{tab:stitching_accuracy}, we see that with unsupervised alignment, we manage to achieve stitching accuracies that are similar to those achieved by using all 1024 points as anchors and computing an orthogonal Procrustes fit (the fully supervised baseline approach).
We hypothesize that this shows for each dataset, the geometry of samples embedded in latent space contains sufficient information to recover an alignment between them.

\begin{table*}
    \centering
    \caption{Stitching accuracy (\%) of fully supervised and unsupervised methods on test sets for each dataset. HGA optimization is done with 50 restarts for each dataset, and the rotation matrix with the highest objective function score is used for model stitching. Results show mean and standard deviation for 3 runs of Latent Functional Maps and HGA with different random seeds.}
    \label{tab:stitching_accuracy}
    \begin{tabular}{@{}l|cc|cc@{}}
        \toprule
        Dataset & Supervised Ortho & Unsupervised Latent Functional Maps & Unsupervised HGA \\
        \midrule
        MNIST    & $96.79\%$ & $27.73 \pm 14.39 \%$ & $96.55 \pm 0.09 \%$ \\
        CIFAR10  & $73.48\%$ & $19.62 \pm  9.54 \%$ & $72.80 \pm 0.30 \%$ \\
        CIFAR100 & $34.51\%$ & $ 1.42 \pm  0.33 \%$ & $31.20 \pm 0.96 \%$ \\
        \bottomrule
    \end{tabular}
\end{table*}

\begin{wrapfigure}{r}{0.5\linewidth}
    \centering
    \vspace{-3mm}
    \includegraphics[width=1.0\linewidth]{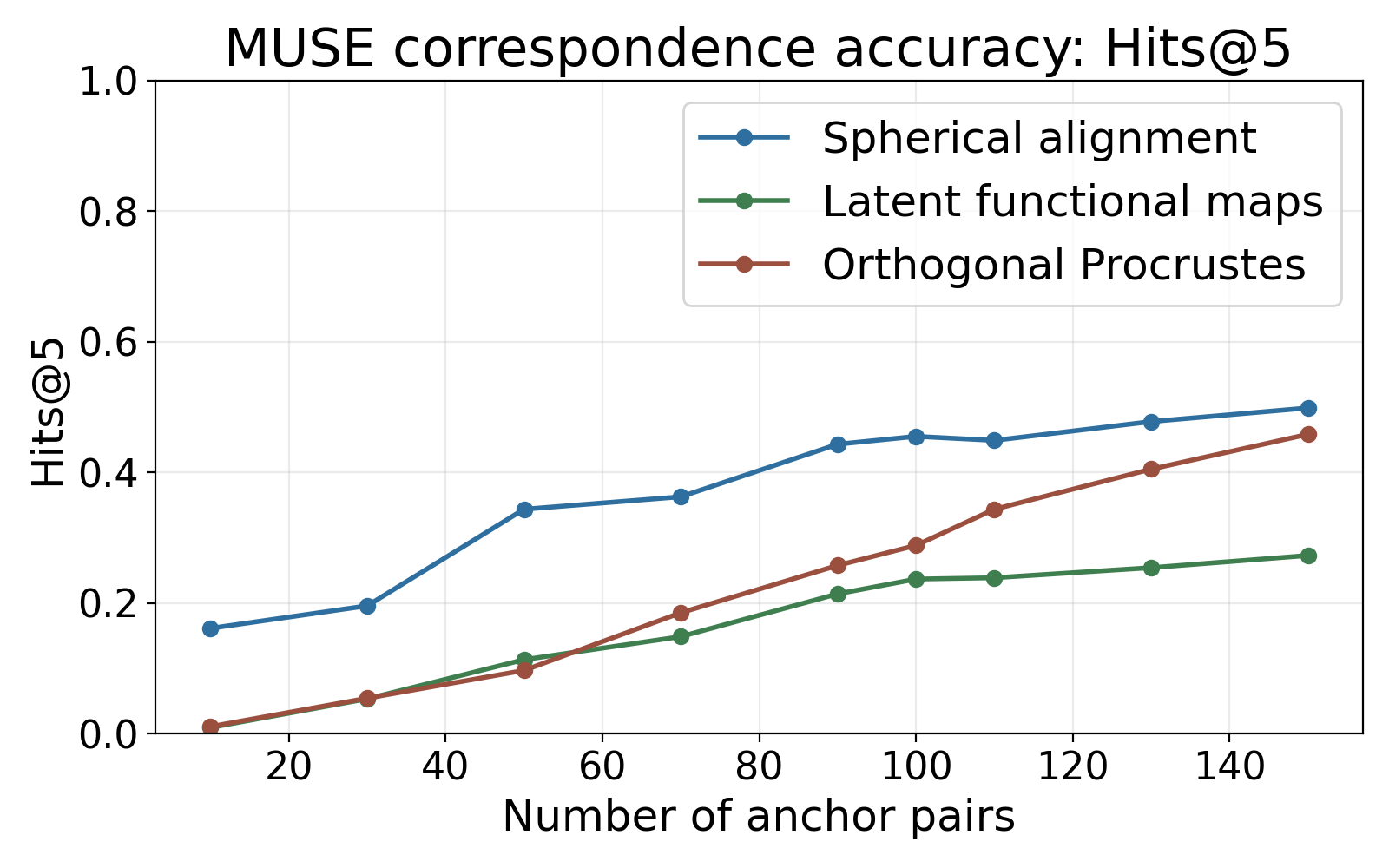}
    \caption{Word correspondence accuracy vs number of anchors for various alignment methods on word embeddings from FastText}
    \vspace{-3mm}
    \label{fig:fasttext_top5}
\end{wrapfigure}

\noindent\textbf{FastText Word Embedding Translation:}
In order to test HGA in the domain of word embeddings, we used FastText's pretrained vectors from the Common Crawl corpus \cite{bojanowski2017enriching, grave2018learningwordvectors157}.
Given two sets of common words from English and French, we manually created a small set of corresponding word pairs from both languages.
Using these word correspondences as anchors, we tested multiple methods of alignment between the two embedding spaces. 
The accuracy of each method was calculated using known correspondences from the MUSE dictionaries \cite{conneau2018wordtranslationparalleldata}. 
We find that in the setting of extremely few anchors, HGA manages much better at finding word correspondences, as seen in Figure \ref{fig:fasttext_top5}.
Qualitatively, we can see how this works in Figure \ref{fig:hga_vs_lfm_qualitative_fasttext}.
While other methods can effectively align the latent space for particular words, HGA manages to use anchor labels as well as geometric information about the two latent spaces in order to translate words that are not anchored. 

\section{Discussion and Conclusion}
HGA aligns independently trained latent spaces by fitting an orthogonal transformation that maximizes a geodesic-kernel measure of geometric fit, matching supervised alignment methods with little or no supervision.
The primary challenge of the proposed method is that the optimization can be unstable due to the large initial bandwidth creating incorrect distribution matches that lead to false alignment.
Fortunately, this problem can be partially addressed by adapting a restart heuristic during the optimization process,
as our experiments (see appendix Figure \ref{fig:objective_function_vs_stitching_accuracy}) have shown that high objective function scores correspond to desirable alignment.
Therefore while it may not always be possible to quickly find an alignment between spaces, it is possible to verify whether an alignment is desirable, and if not keep trying until a desirable one is found.
As the latent spaces that it is applied to become increasingly complex, it becomes more difficult to find useful maxima of the objective function $J_\sigma (R)$.
For model stitching on simple datasets such as MNIST or CIFAR10 we can easily find an optimal transformation between them with stitching accuracy similar to fully supervised approaches.
For more complex latent spaces such as the CLIP embedding space, or FastText word embeddings, anchors are often still needed obtain useful results.
Overall, we find that HGA is an effective method for latent space alignment.
We hypothesize that the geometries of different latent spaces possess signatures that are unique up to the data they represent, and their usefulness for alignment depends on how little symmetry is present.

Future work will be organized as follows:
 (1) Improving the algorithm for optimization by leveraging more geometric understanding of latent spaces.
 Currently it requires about an hour to find a decent alignment between two spaces.
 Further work could parallelize restarts and further develop the early stop heuristic to improve computation time.
 We wish to maximize the chances of finding a desirable objective function maxima, and quantify how difficult this will be for various latent spaces.
 (2) Characterizing latent spaces based on concrete geometric information.
 We seek to find a distinct way to describe latent spaces that is invariant to certain basic transformations.
 This could be used to quantify the ability of latent spaces to be aligned.

\begin{ack}
This work was performed under the auspices of the U.S.\ Department of Energy (DOE) by Lawrence Livermore National Laboratory under Contract No. DE-AC52-07NA27344. This work was supported by DOE ECRP 51917/SCW1885. 
This work is reviewed and released under LLNL-PROC-2023638.
\end{ack}

\bibliographystyle{unsrt}
\bibliography{references}
\medskip

\appendix
\section{Optimization Method}
\subsection{Overview}

Optimization is performed by starting from a random orthogonal matrix, and iteratively improving it using gradient ascent. 
We calculate gradients by projecting the Euclidean gradient into the tangent space of $O(d)$.
Then an update is performed using the Adam optimizer \cite{Kingma:2014vow} and the resulting $R$ is mapped back into $O(d)$ using Singular Value Decomposition.
We also explored updating according to the Euclidean gradient and mapping the result back into $O(d)$, however this was found to be much less effective. 
We use a continuation strategy during the optimization process by starting with a large $\sigma$, and progressively decreasing it.

The process is typically restarted with different random initializations until an $R$ matrix satisfying a specified threshold for $J_\sigma (R)$ is found. Alternatively, a fixed number of restarts can be used, retaining the solution with the highest score. Our experiments indicate that multiple restarts improve the likelihood of finding a desirable maximum of the objective function.

\subsection{Riemannian Gradient}
\label{sec:gradient}
The Euclidean gradient can, and often will point away from the constraint set:
\[
G = \nabla_R J_\sigma (R)
\]
Therefore, for the best optimization on $O(d)$, we project it into the tangent space of $O(d)$:
\[
\text{grad} J_\sigma(R) = G - R\,\text{sym}(R^TG)
\]
Where: 
\[\text{sym}(A)=\frac{1}{2}\big(A + A^T\big)\]
Then, in order to move towards optimizing our objective, we take a a positive gradient step, and retract on to the orthonormal group: 
\[\tilde{R} = R + \eta \,\text{grad}\,J_\sigma(R)\]
\[\tilde{R} = U\Sigma V^T\to R_{\text{new}} = UV^T\]

\subsection{Bandwidth Continuation}
Instead of optimizing a single kernel, we lower the $\sigma$ hyperparameter, solving a sequence: 
\[\sigma_1>\sigma_2>\cdots>\sigma_L\]
And at each stage, we initialize from the previous solution. Intuitively, at large $\sigma$, the objective function $J_\sigma$ will be highly smooth, considering many pairs and will be nearly concave. At smaller $\sigma$, it will only consider local nearby pairs and therefore be highly nonconcave. We can think of this process as a continuation from global distribution alignment, to cluster alignment, to local point alignment.

\subsection{Learning Rate Scheduling}
Our optimization algorithm uses a cosine annealing decay.
Given hyperparameters $\eta_\text{min}$ and $\eta_\text{max}$ we decay according to: 
\[
\eta_t =
\eta_\text{min} +
\frac{1}{2} (\eta_\text{max}-\eta_\text{min})\bigg(1 + \cos \frac{\pi t}{T} \bigg)
\]

\subsection{Early Stop Heuristic}
Due to the nature of the problem, many restarts are needed to obtain useful results, because many runs will converge to undesirable local minima.
This is because when the kernel bandwidth is large at the beginning of the optimization, many potential symmetries exist, however they often lead to an incorrect alignment.
As the kernel bandwidth is decreased, the symmetries which existed at a high bandwidth may cease to be present, causing the optimization to be stuck at an undesirable local maxima. 
Therefore we find it necessary to restart the optimization algorithm many times to increase the chances of finding a desirable objective function maxima. \\

\begin{figure}
    \centering
    \includegraphics[width=1.0\linewidth]{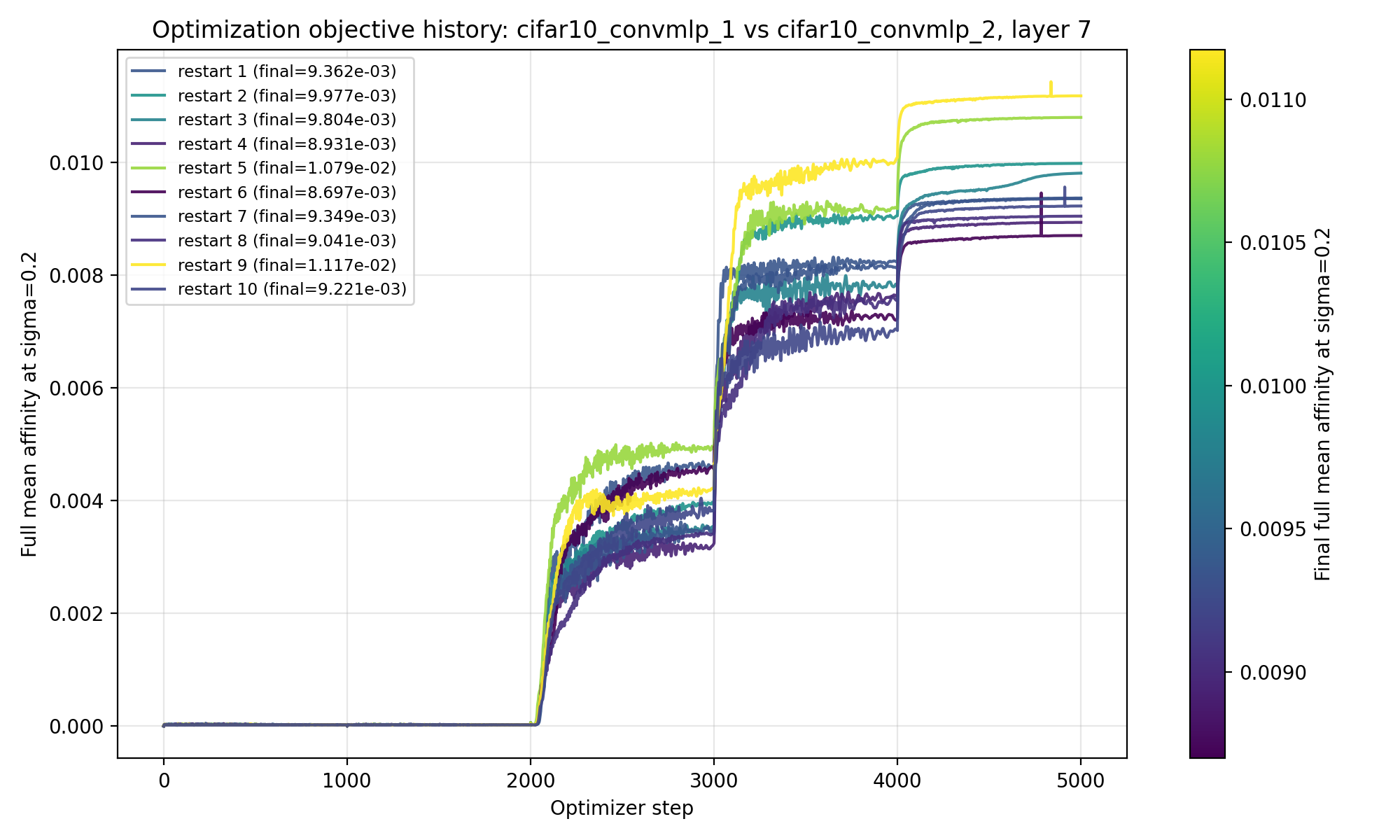}
    \caption{Objective function score at $\sigma=0.2$ vs optimization step for 10 random orthogonal matrix initializations. Color denotes the final objective function score reached. Noticeable jumps are due to a decrease in $\sigma$ for the objective function $J_\sigma(R)$ being used to compute optimization gradient.}
    \label{fig:objective_function_tracking}
\end{figure}

Due to the computational resources required to perform this across many restarts, we investigated the possibility of doing early stops during the optimization process in order to end runs with lower chances of finding a desirable maxima. 
By observing Figure \ref{fig:objective_function_tracking}, we see that in by the mid stage, it is not possible to know exactly which run will be the maximum, however it is for later stages. 
In some experiments, we determined an threshold for the final stage, and stopped restarts below the threshold at the beginning of the final stage.

\section{Hyperparameter Experiment}
While we have found useful results from HGA, it is still highly unstable.
Often, small tweaks to hyperparameters have inexplicably large effects on the outcome of the alignment task (eg. model stitching accuracy).
We performed a study on the hyperparameters of HGA for performing model stitching on every layer of a model trained on CIFAR10.
For each of the following hyperparameters, we tried tweaking them in several variations and viewing the impact that this had on the model stitching accuracy. 
After this experiment was performed, we record the hyperparameter configurations with the best stitching accuracies in Table \ref{tab:cifar10_best_stitching_accuracies}.
While the experiment was unable to find a way to make HGA stable, it did reveal an ideal configuration of hyperparameters, and yielded stitching accuracy comparable to their fully supervised counterparts for each layer.
We find that it is optimal to start with a large $\sigma$ of around $4$, and work down to a smaller $\sigma$ of around $0.1$.

\begin{table}[]
    \centering
    \begin{tabular}{ccccccc}
        \toprule
        Layer & $\sigma$ & $\sigma_\text{start}$ & LR & LR final & \makecell{Stitching Acc.\\ (HGA unsupervised)} & \makecell{Stitching Acc.\\ (Ortho Supervised)} \\
        \midrule
        4 & 0.1 & 4 & 0.1  & 0.0005 & 70\% & 70\% \\
        5 & 0.1 & 4 & 0.1  & 0.0005 & 62\% & 71\% \\
        6 & 0.1 & 4 & 0.1  & 0.001  & 75\% & 73\% \\
        7 & 0.2 & 4 & 0.01 & 0.0005 & 75\% & 76\% \\
        8 & 0.1 & 4 & 0.01 & 0.0005 & 76\% & 77\% \\
        \bottomrule
        \hspace{1.0pt}
    \end{tabular}
    \caption{CIFAR10 best unsupervised stitching accuracies and corresponding hyperparameters for each model layer. Results from an experiment running the optimization algorithm with a single restart and different hyperparameters for each run. Far right column shows stitching accuracy for orthogonal least squares solution with 1024 anchor points.}
    \label{tab:cifar10_best_stitching_accuracies}
\end{table}

\subsection{Anchor Corruption Test}
An interesting experiment to showcase the properties of the HGA method is observing how robust alignment methods are to corrupted anchors. 
Given latent space samples from two neural networks, we progressively swapped out anchors with correspondences to points from incorrect classes. 
For each alignment method, we observe how this effects the stitching accuracy.
We find that in the latent spaces of two neural networks trained on the MNIST dataset, HGA is able to have a nearly unchanged stitching accuracy, even with up to $85\%$ of its anchors corrupted. 
\begin{figure}
    \centering
    \includegraphics[width=0.75\linewidth]{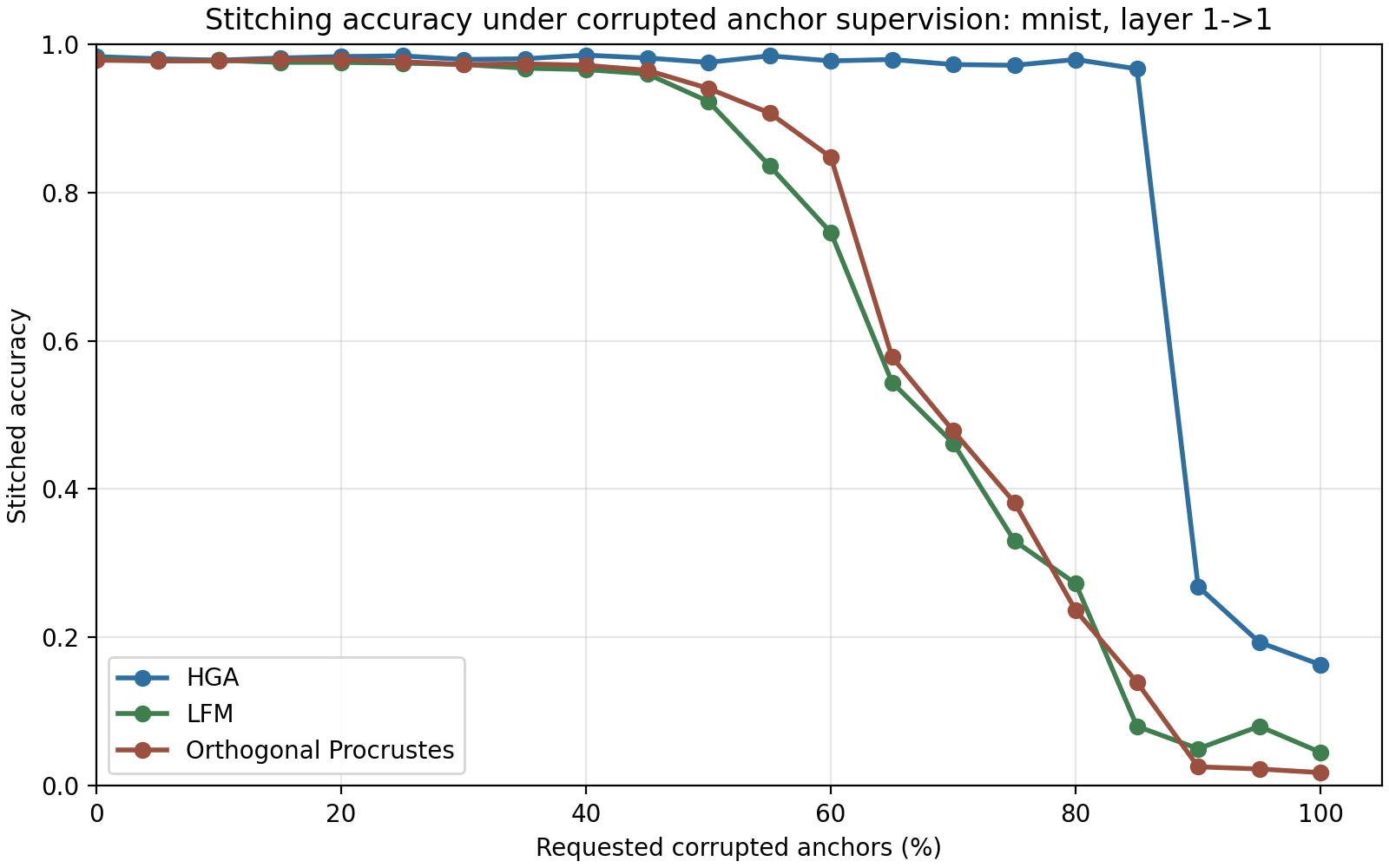}
    \caption{Percent of Anchors Corrupted vs. Model Stitching Accuracy for the latent spaces of two neural networks trained on the MNIST dataset.}
    \label{fig:placeholder}
\end{figure}
\section{Verifying the Usefulness of the Objective Function for Unsupervised Alignment}
\begin{figure}
    \centering
    \begin{subfigure}[b]{0.32\linewidth}
        \includegraphics[width=\linewidth]{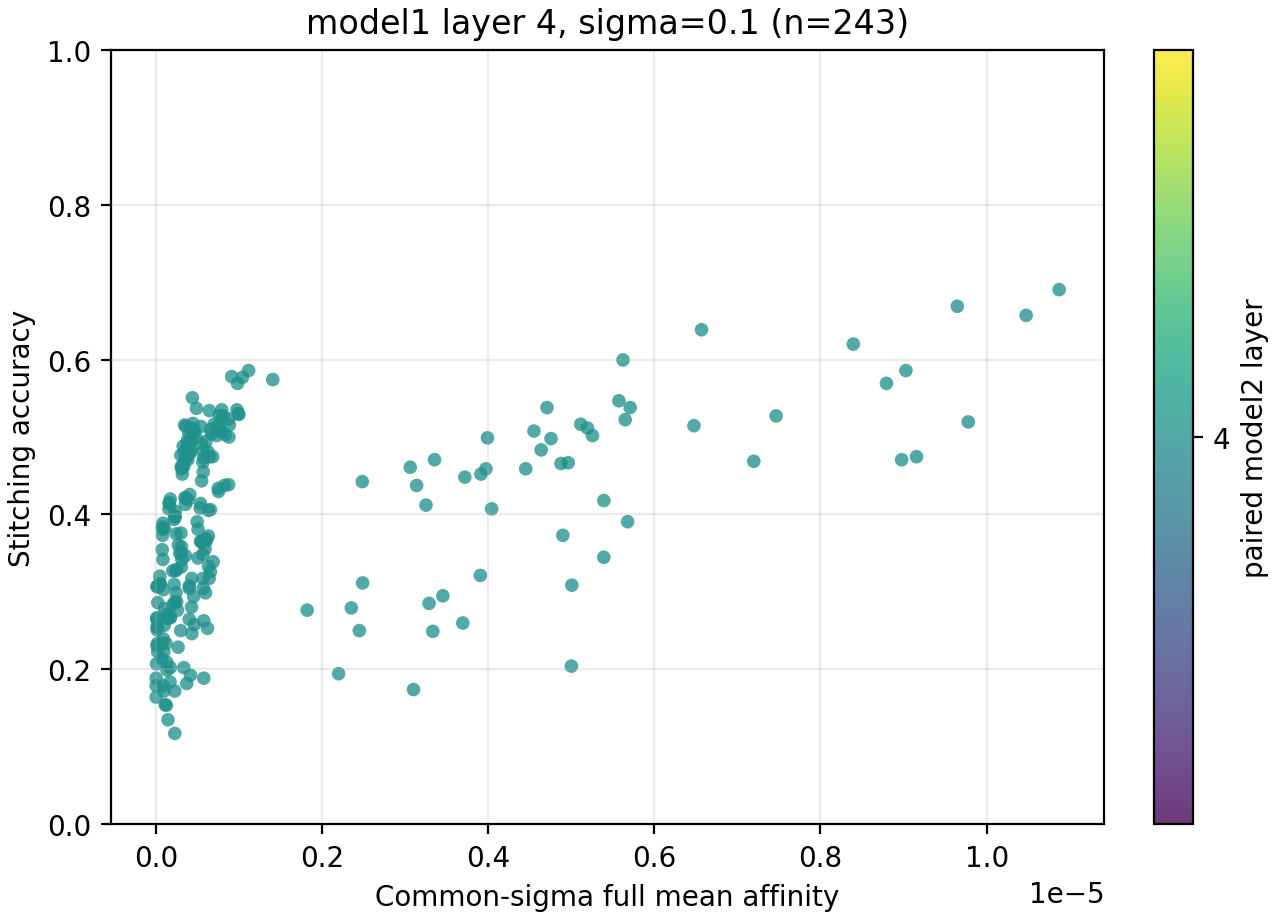}
        \caption{$\sigma=0.1$}
        \label{fig:scatter_1}
    \end{subfigure}
    \hfill
    \begin{subfigure}[b]{0.32\linewidth}
        \includegraphics[width=\linewidth]{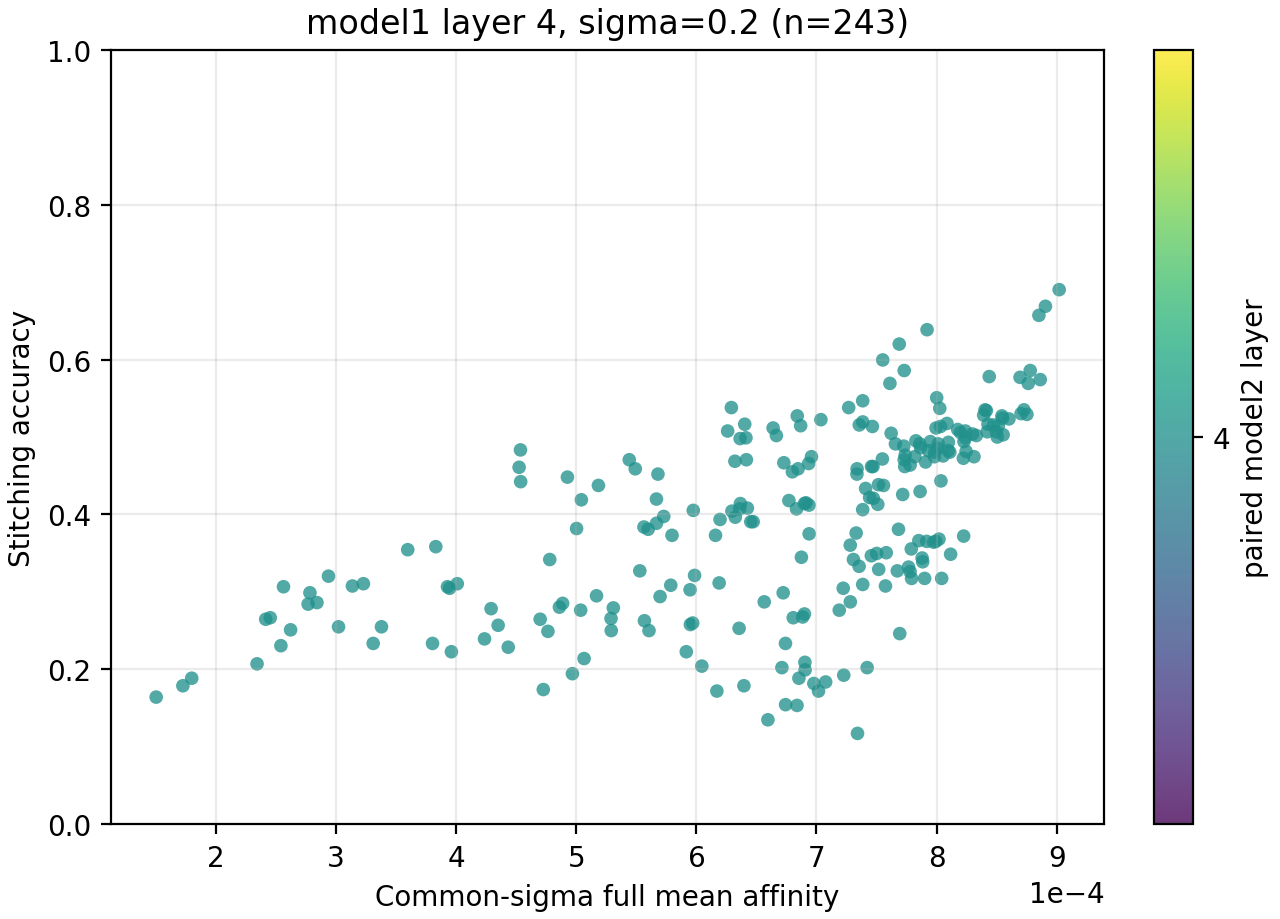}
        \caption{$\sigma=0.2$}
        \label{fig:scatter_2}
    \end{subfigure}
    \hfill
    \begin{subfigure}[b]{0.32\linewidth}
        \includegraphics[width=\linewidth]{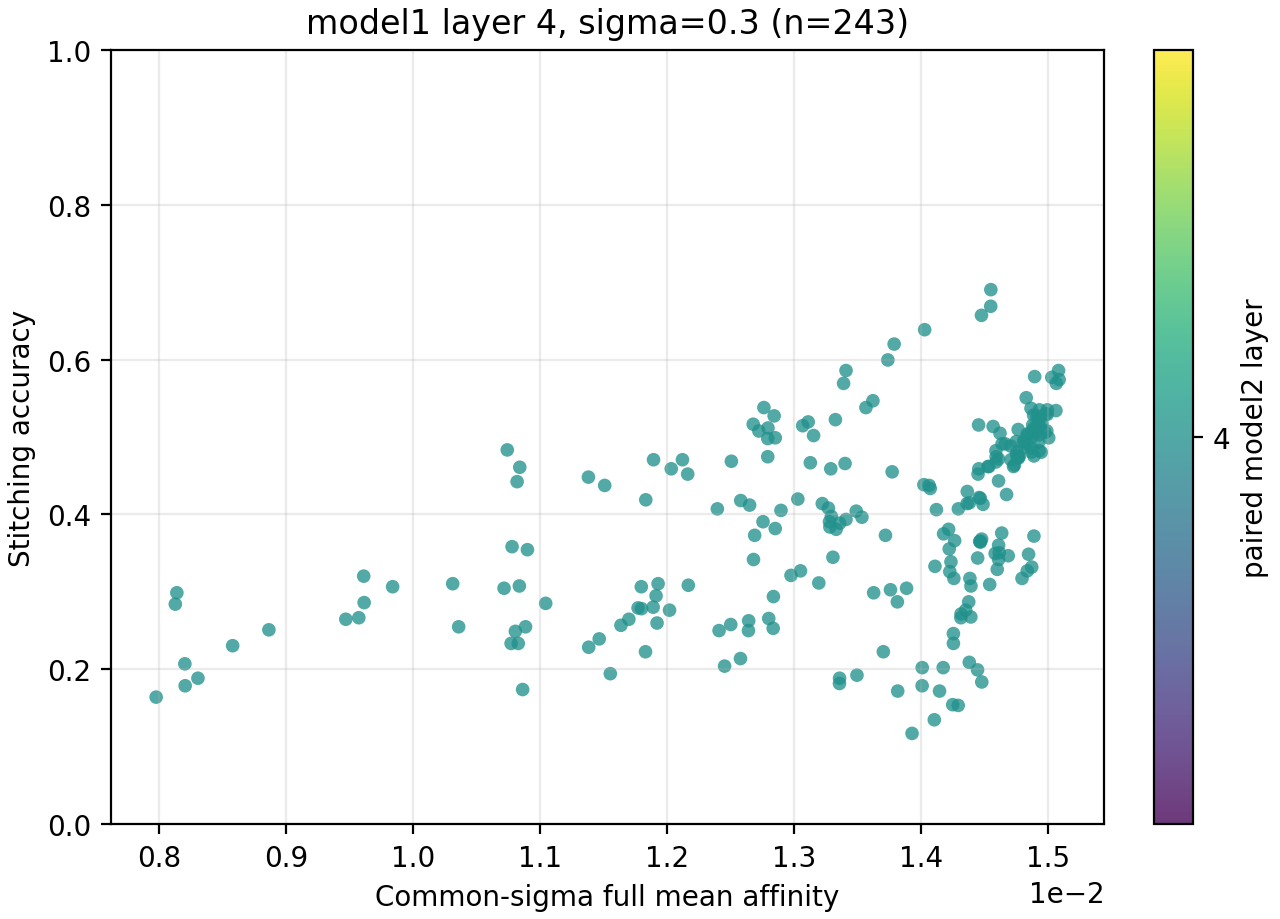}
        \caption{$\sigma=0.3$}
        \label{fig:scatter_3}
    \end{subfigure}

    \vspace{0.5em} 

    \begin{subfigure}[b]{0.32\linewidth}
        \includegraphics[width=\linewidth]{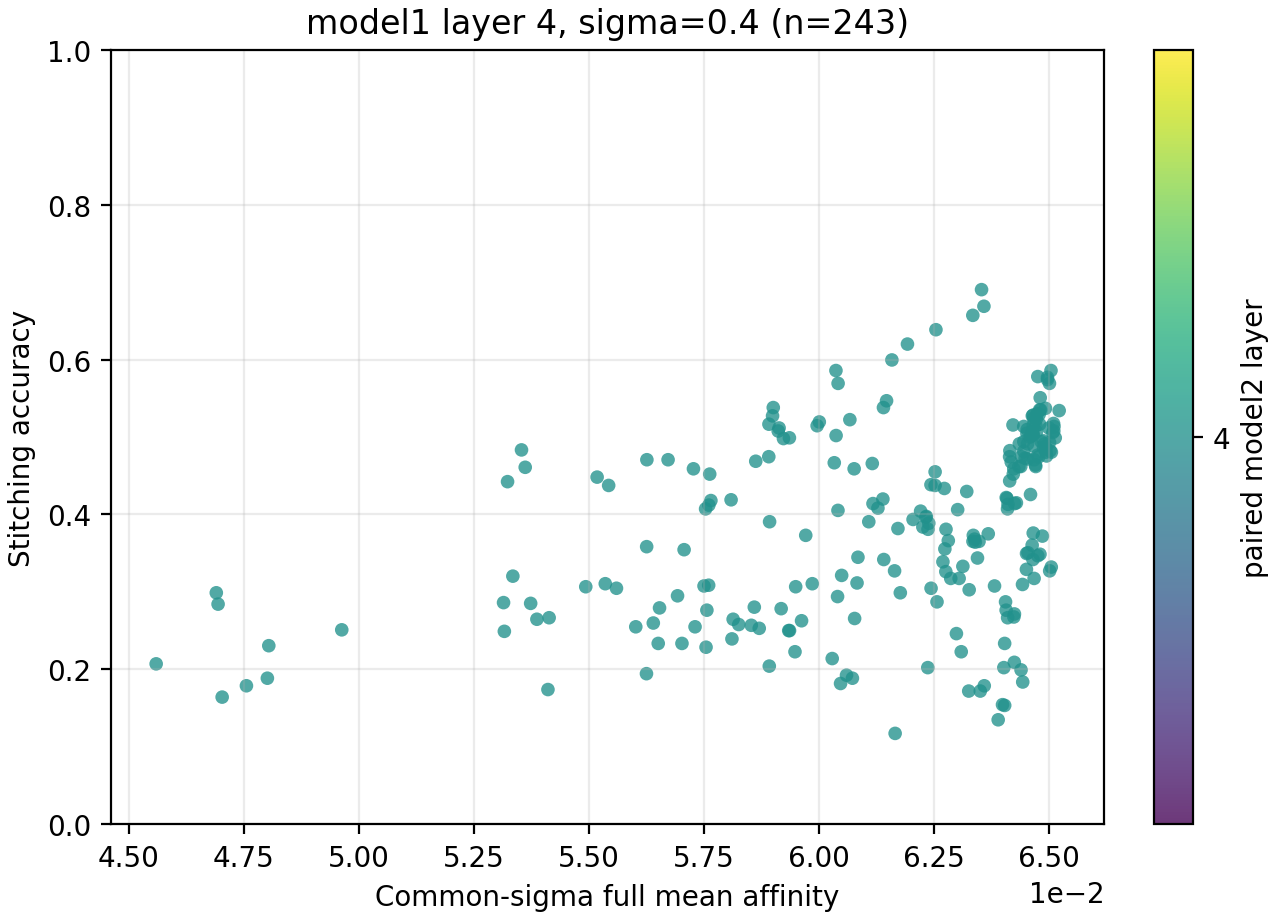}
        \caption{$\sigma=0.4$}
        \label{fig:scatter_4}
    \end{subfigure}
    \hfill
    \begin{subfigure}[b]{0.32\linewidth}
        \includegraphics[width=\linewidth]{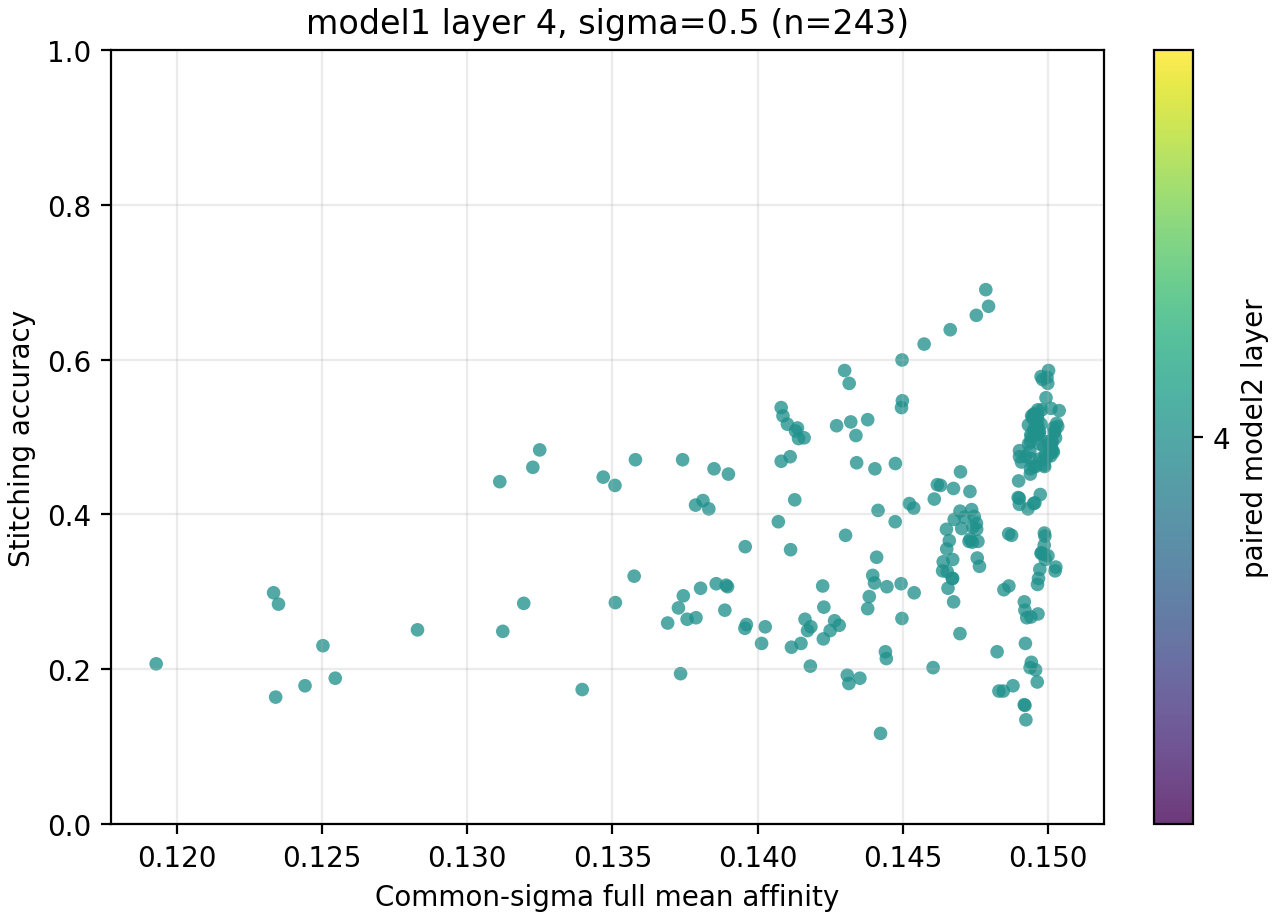}
        \caption{$\sigma=0.5$}
        \label{fig:scatter_5}
    \end{subfigure}
    \hfill
    \begin{subfigure}[b]{0.32\linewidth}
        \includegraphics[width=\linewidth]{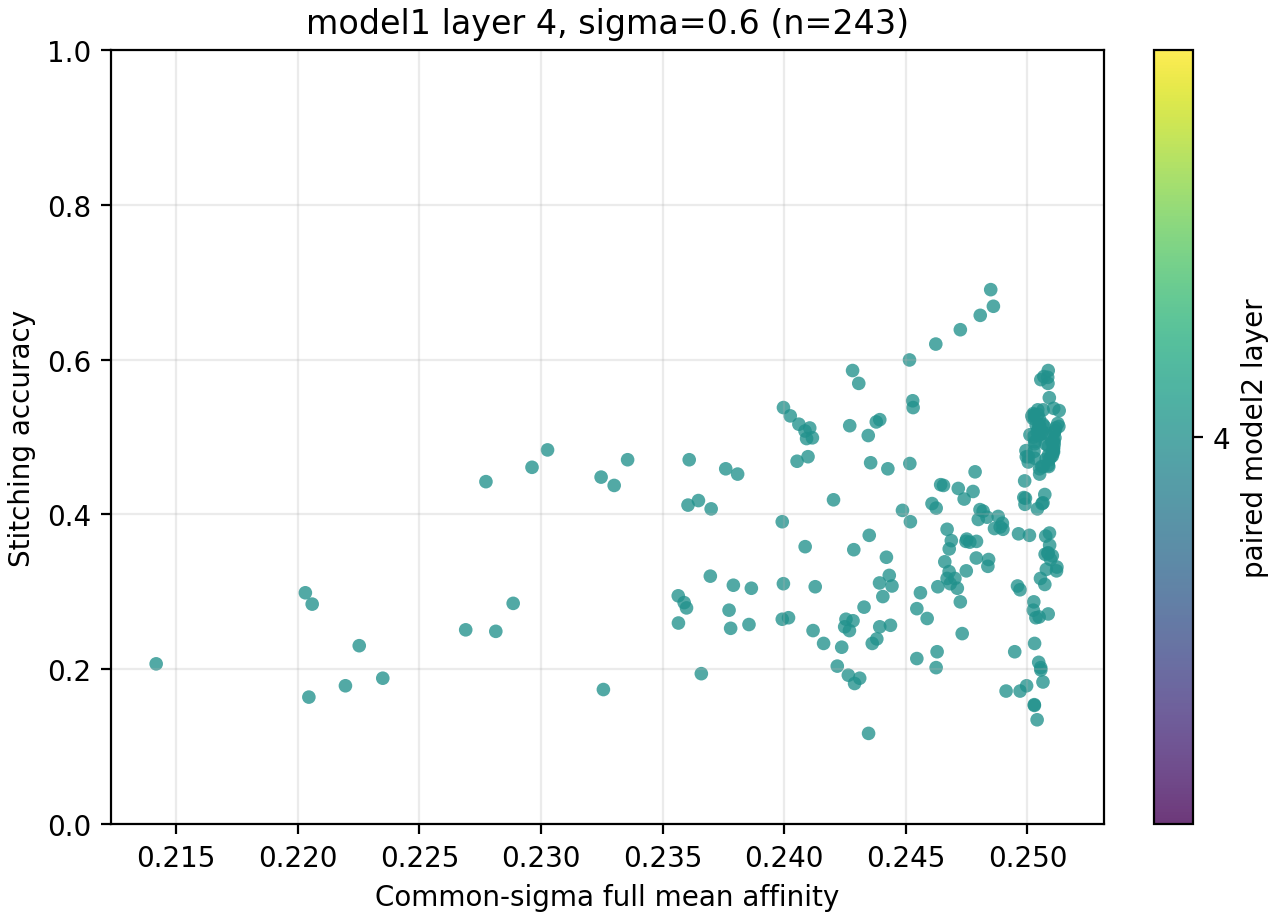}
        \caption{$\sigma=0.6$}
        \label{fig:scatter_6}
    \end{subfigure}

    \caption{Scatter plots show objective function scores vs stitching accuracies for various optimization trials with slightly different optimization hyperparameters on CIFAR10.}
    \label{fig:objective_function_vs_stitching_accuracy}
\end{figure}

Since the algorithm can often be unstable, it is necessary to know whether the objective function maxima that it has discovered is useful in applications which involve unsupervised latent space communication.
Therefore, we wish to validate how well the value of the objective function corresponds to a ``correct fit" between latent spaces being discovered.
We do this by looking at the correlation between the best objective function value, and model stitching accuracy for CIFAR10 and CIFAR100
We performed the optimization for various hyperparameters, and then for each resulting $R$ matrix, computed the objective function $J_\sigma(R)$ at different values of $\sigma$.
The results of this experiment are shown in Figure \ref{fig:objective_function_vs_stitching_accuracy}.
Upon examining the results from this figure, we make two important observations:
\begin{enumerate}
    \item For a value of $\sigma\approx 0.2$, there is a strong correlation between objective function value and stitching accuracy in the regime of high objective function value. 
    \item For larger values of $\sigma$ we see that a high objective function value is necessary but not sufficient for high stitching accuracy. 
\end{enumerate}

\section{Model Stitching Supplemental Information}
\begin{figure}
    \centering
    \begin{subfigure}[b]{0.32\linewidth}
        \includegraphics[width=\linewidth]{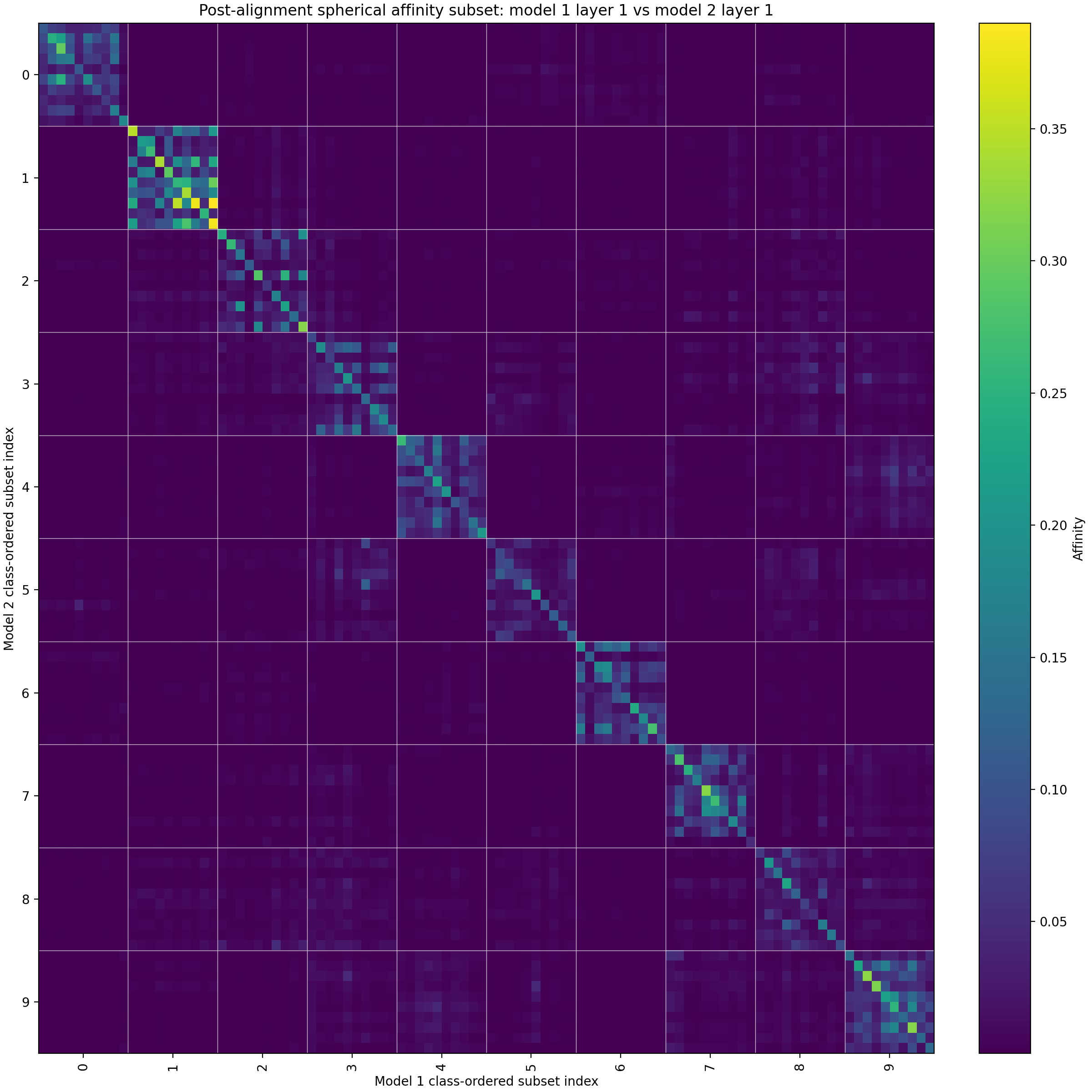}
        \caption{MNIST}
        \label{fig:mnist_kernel}
    \end{subfigure}
    \hfill
    \begin{subfigure}[b]{0.32\linewidth}
        \includegraphics[width=\linewidth]{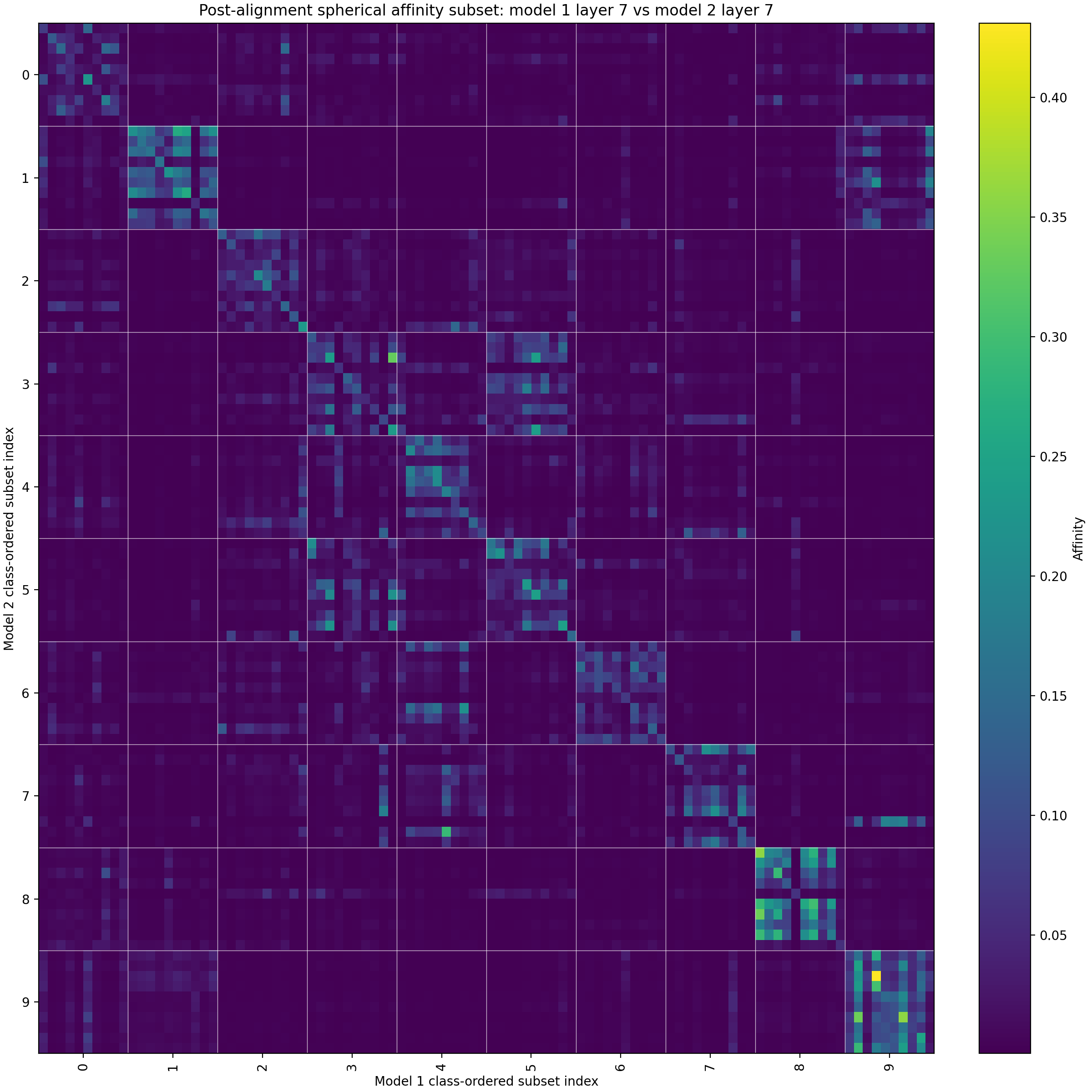}
        \caption{CIFAR10}
        \label{fig:cifar10_kernel}
    \end{subfigure}
    \hfill
    \begin{subfigure}[b]{0.32\linewidth}
        \includegraphics[width=\linewidth]{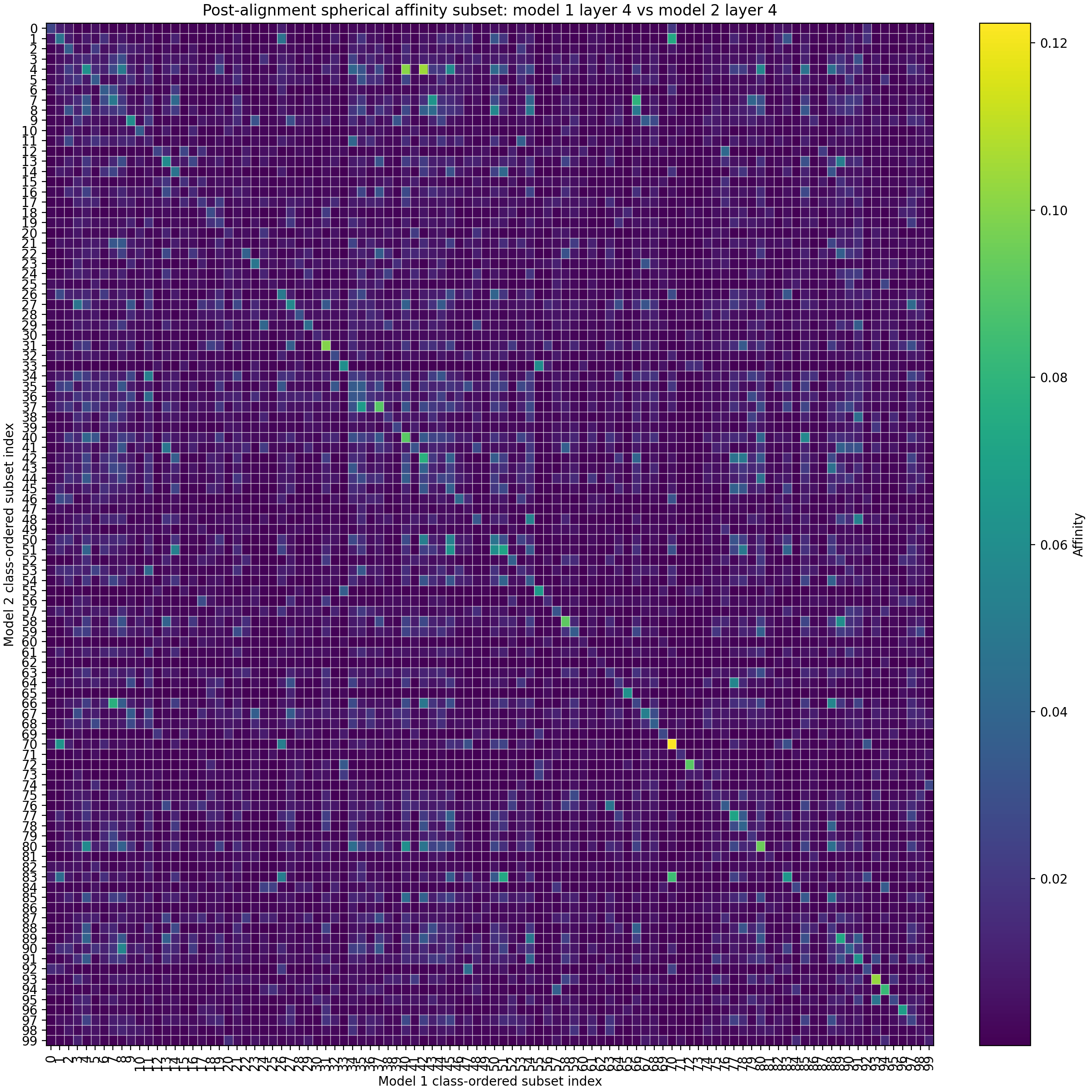}
        \caption{CIFAR100}
        \label{fig:cifar100_kernel}
    \end{subfigure}

    \vspace{0.5em} 

    \begin{subfigure}[b]{0.32\linewidth}
        \includegraphics[width=\linewidth]{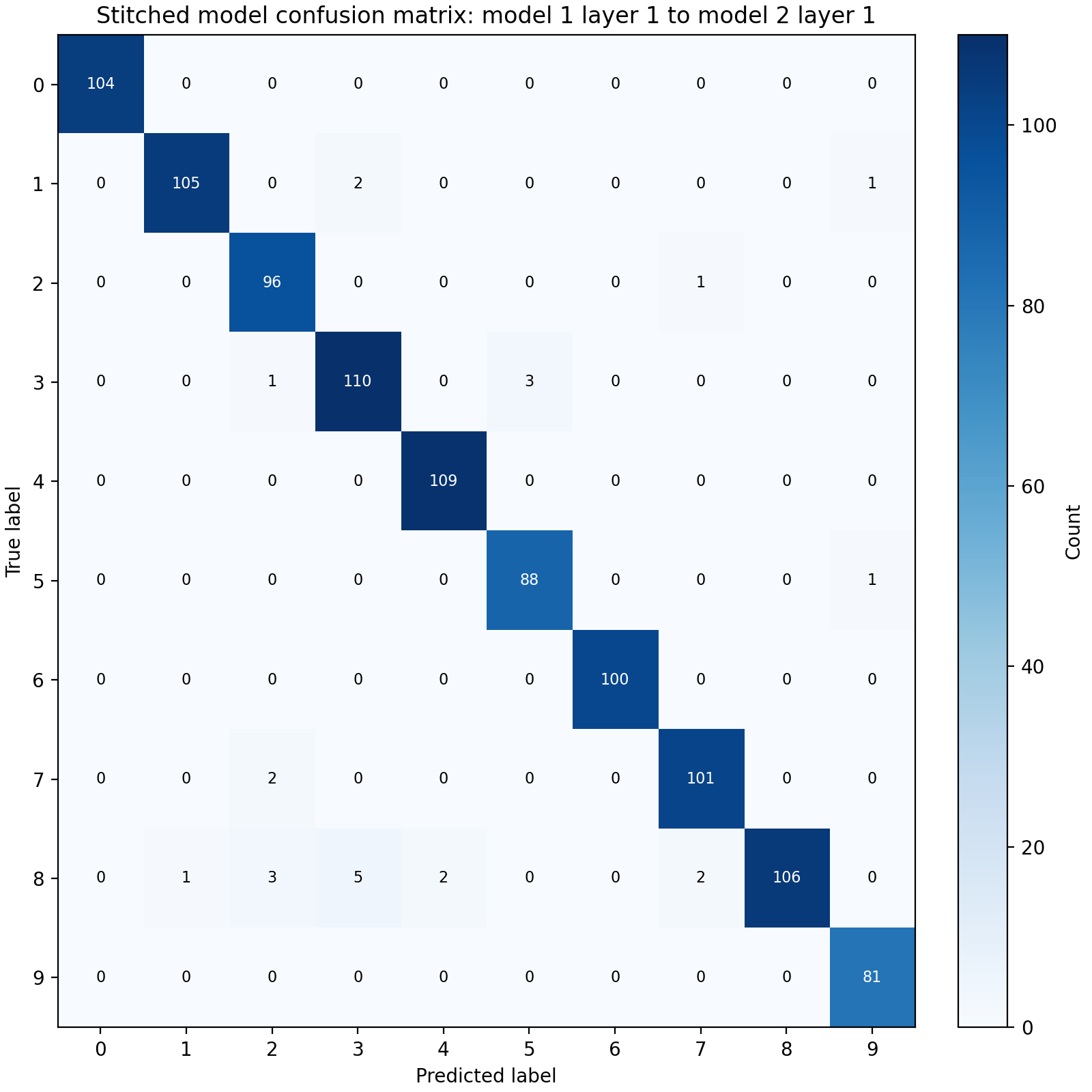}
        \caption{MNIST}
        \label{fig:mnist_confusion}
    \end{subfigure}
    \hfill
    \begin{subfigure}[b]{0.32\linewidth}
        \includegraphics[width=\linewidth]{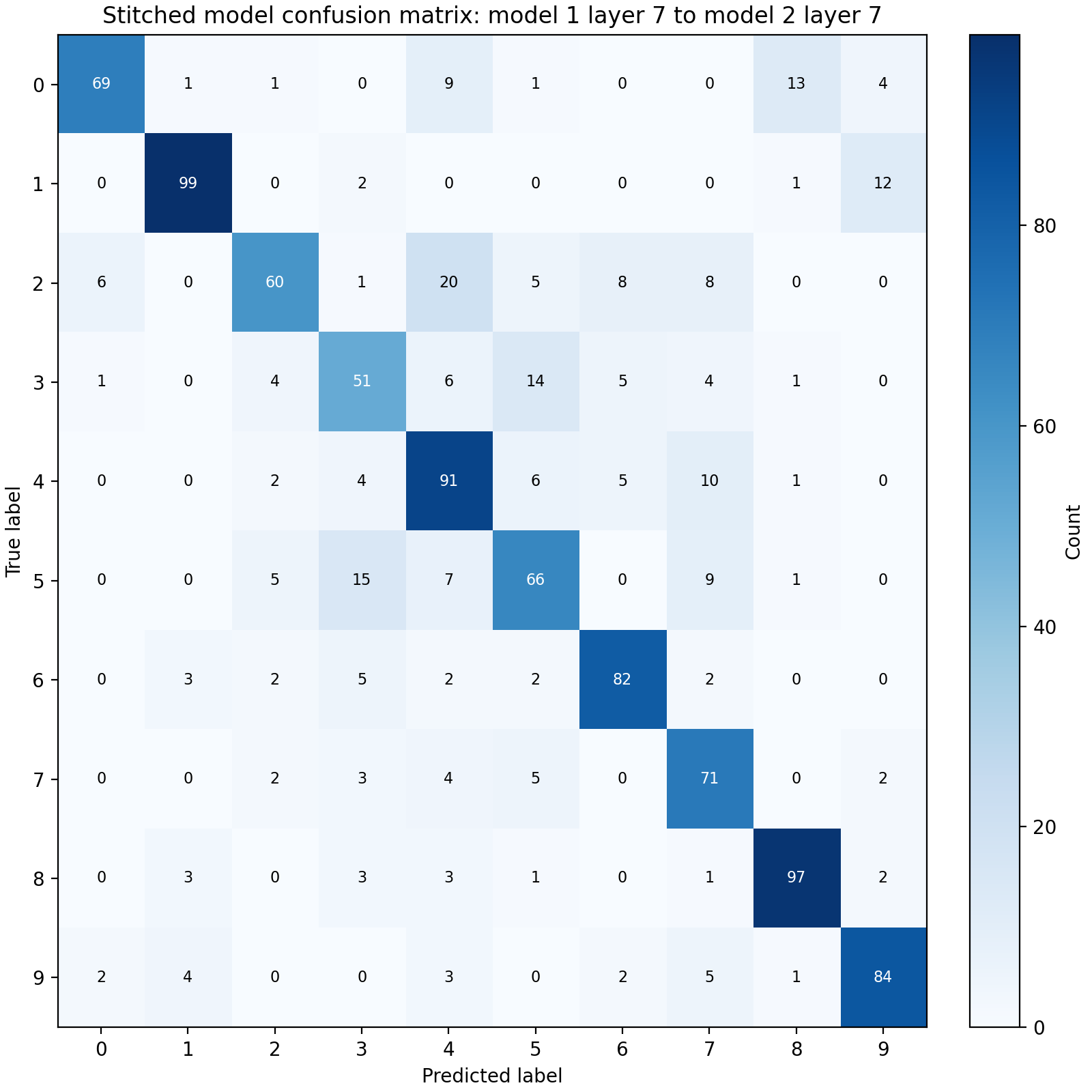}
        \caption{CIFAR10}
        \label{fig:cifar10_confusion}
    \end{subfigure}
    \hfill
    \begin{subfigure}[b]{0.32\linewidth}
        \includegraphics[width=\linewidth]{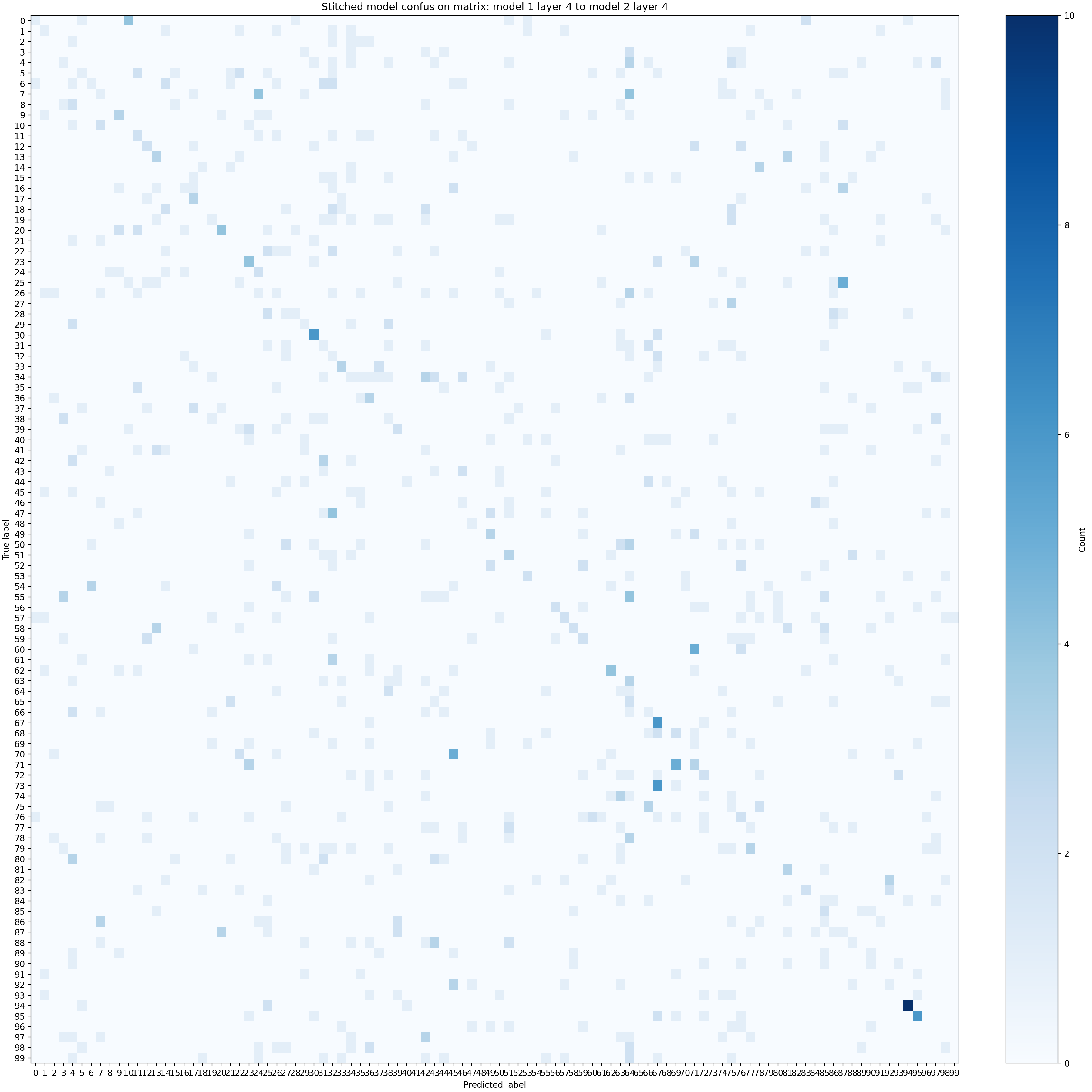}
        \caption{CIFAR100}
        \label{fig:cifar100_confusion}
    \end{subfigure}

    \caption{Implied kernel subsets and model stitching confusion matrices for unsupervised stitching in MNIST, CIFAR10, and CIFAR100}
    \label{fig:stitching_kernel_and_confusion_matrix}
\end{figure}

Model stitching is an important task to demonstrate the effectiveness of HGA at unsupervised latent space communication.
After using the optimization algorithm to find a desirable $R$ matrix, we wish to see how well it can find correspondences between points, and classes between the two latent spaces.
In order to compare points between the different latent spaces, we define a kernel function: $k:S_\M\to S_\N$ by: 
\[k(x,y)=\exp\Bigg(-\frac{|Rx-y|^2}{2\sigma^2}\Bigg)\]
By examining the top row of Figure \ref{fig:stitching_kernel_and_confusion_matrix}, we see the kernel values for a random subset from each dataset, organized by class.
We can qualitatively observe that the kernel is adept at recovering the correspondences between points from their respective latent space positions.

\section{Qualitative Results for FastText Alignment}
Finding alignment between FastText word embeddings is a useful example to show how HGA can use the anchors that it has been provided, and make new inferences about how the Latent Spaces can be aligned. 
In figure \ref{fig:hga_vs_lfm_qualitative_fasttext} we can see this demonstrated.
In this example, 100 anchor words are given to the HGA and LFM \cite{fumero2024latent} methods.
While both are able to effectively represent known correspondences between the same words, we see that HGA is also able to infer correspondences between words which are not in the anchor list. 
\begin{figure*}
    \centering
    \scriptsize
    \setlength{\tabcolsep}{3pt}
    \begin{minipage}[t]{0.48\linewidth}
    \centering
    \begin{tabular}{@{}rlr@{\hspace{1.5em}}rlr@{}}
        \toprule
        \multicolumn{3}{c}{\textbf{LFM}} & \multicolumn{3}{c}{\textbf{HGA}} \\
        \cmidrule(lr){1-3} \cmidrule(lr){4-6}
        R & Word & Score & R & Word & Score \\
        \midrule
        \multicolumn{6}{c}{\textit{dog} (in anchor list)} \\
        1 & \textbf{chien} & 0.050 & 1 & \textbf{chien} & 0.929 \\
        2 & chat & 0.040 & 2 & chiens & 0.692 \\
        3 & animaux & 0.031 & 3 & chat & 0.683 \\
        4 & peau & 0.028 & 4 & animal & 0.609 \\
        5 & espèce & 0.021 & 5 & chats & 0.583 \\
        6 & poisson & 0.021 & 6 & cheval & 0.513 \\
        7 & bébé & 0.021 & 7 & animaux & 0.452 \\
        8 & enfant & 0.018 & 8 & bébé & 0.443 \\
        9 & homme & 0.018 & 9 & enfant & 0.399 \\
        10 & espèces & 0.017 & 10 & maman & 0.389 \\
        \midrule
        \multicolumn{6}{c}{\textit{house} (in anchor list)} \\
        1 & \textbf{maison} & 0.046 & 1 & \textbf{maison} & 0.795 \\
        2 & maisons & 0.040 & 2 & villa & 0.541 \\
        3 & Maison & 0.035 & 3 & jardin & 0.536 \\
        4 & appartement & 0.030 & 4 & garage & 0.516 \\
        5 & immobilier & 0.027 & 5 & appartement & 0.505 \\
        6 & terrain & 0.023 & 6 & maisons & 0.492 \\
        7 & vente & 0.021 & 7 & Maison & 0.483 \\
        8 & pierre & 0.020 & 8 & chambre & 0.481 \\
        9 & village & 0.020 & 9 & immeuble & 0.468 \\
        10 & jardin & 0.020 & 10 & château & 0.458 \\
        \midrule
        \multicolumn{6}{c}{\textit{bird} (not in anchor list)} \\
        1 & Posté & 0.041 & 1 & poisson & 0.447 \\
        2 & poisson & 0.030 & 2 & poulet & 0.437 \\
        3 & ciel & 0.030 & 3 & chat & 0.432 \\
        4 & Fichier & 0.027 & 4 & arbre & 0.389 \\
        5 & restaurant & 0.025 & 5 & chien & 0.383 \\
        6 & \dag & 0.025 & 6 & animal & 0.371 \\
        7 & vitesse & 0.025 & 7 & \textbf{oiseau} & 0.366 \\
        8 & Institut & 0.025 & 8 & pomme & 0.341 \\
        9 & pont & 0.024 & 9 & bateau & 0.332 \\
        10 & Service & 0.023 & 10 & jardin & 0.320 \\
        \bottomrule
    \end{tabular}
    \end{minipage}
    \hfill
    \begin{minipage}[t]{0.48\linewidth}
    \centering
    \begin{tabular}{@{}rlr@{\hspace{1.5em}}rlr@{}}
        \toprule
        \multicolumn{3}{c}{\textbf{LFM}} & \multicolumn{3}{c}{\textbf{HGA}} \\
        \cmidrule(lr){1-3} \cmidrule(lr){4-6}
        R & Word & Score & R & Word & Score \\
        \midrule
        \multicolumn{6}{c}{\textit{horse} (not in anchor list)} \\
        1 & chien & 0.037 & 1 & chien & 0.480 \\
        2 & chat & 0.032 & 2 & \textbf{cheval} & 0.416 \\
        3 & voiture & 0.031 & 3 & pied & 0.396 \\
        4 & John & 0.026 & 4 & animal & 0.375 \\
        5 & Livraison & 0.026 & 5 & chat & 0.359 \\
        6 & animaux & 0.026 & 6 & chiens & 0.349 \\
        7 & peau & 0.025 & 7 & homme & 0.342 \\
        8 & compagnie & 0.025 & 8 & poulet & 0.315 \\
        9 & police & 0.025 & 9 & marche & 0.309 \\
        10 & russe & 0.024 & 10 & pieds & 0.309 \\
        \midrule
        \multicolumn{6}{c}{\textit{flower} (not in anchor list)} \\
        1 & arbre & 0.041 & 1 & jardin & 0.519 \\
        2 & étoile & 0.041 & 2 & arbre & 0.448 \\
        3 & lit & 0.039 & 3 & \textbf{fleur} & 0.407 \\
        4 & TV & 0.035 & 4 & motif & 0.387 \\
        5 & argent & 0.034 & 5 & vert & 0.387 \\
        6 & étoiles & 0.032 & 6 & arbres & 0.377 \\
        7 & or & 0.031 & 7 & thé & 0.374 \\
        8 & importance & 0.031 & 8 & rose & 0.370 \\
        9 & bébé & 0.030 & 9 & plante & 0.370 \\
        10 & amour & 0.029 & 10 & coton & 0.366 \\
        \midrule
        \multicolumn{6}{c}{\textit{forest} (not in anchor list)} \\
        1 & vert & 0.046 & 1 & \textbf{forêt} & 0.495 \\
        2 & rivière & 0.040 & 2 & arbre & 0.453 \\
        3 & bleu & 0.034 & 3 & forêts & 0.410 \\
        4 & \textbf{forêt} & 0.031 & 4 & montagne & 0.406 \\
        5 & zone & 0.031 & 5 & arbres & 0.392 \\
        6 & Posté & 0.031 & 6 & village & 0.385 \\
        7 & orange & 0.031 & 7 & sauvage & 0.375 \\
        8 & sources & 0.028 & 8 & paysage & 0.374 \\
        9 & Date & 0.027 & 9 & naturel & 0.373 \\
        10 & '' & 0.027 & 10 & jardin & 0.363 \\
        \bottomrule
    \end{tabular}
    \end{minipage}

    \caption{Qualitative comparison of nearest French neighbours retrieved by LFM and HGA for English query words, given 100 anchor words. Correct translations are shown in \textbf{bold}. For anchor words (\textit{dog}, \textit{house}) both methods rank the correct translation first, but for words absent from the anchor list (\textit{bird}, \textit{horse}, \textit{flower}, \textit{forest}) LFM largely returns unrelated or noisy tokens, whereas HGA still recovers the correct translation.}
    \label{fig:hga_vs_lfm_qualitative_fasttext}
\end{figure*}
\section{Experiments with Distribution Matching}
As part of this work, it is relevant to investigate whether or not separate latent spaces match.
We performed two small experiments investigating how well we can locate a mismatch between latent spaces.
Formally, suppose that we have two spaces $X$ and $Y$, a transformation between them $T:X\to Y$ and a kernel function $k:X\times Y \to \mathbb{R}^+$. For this experiment, we use the Gaussian Kernel between the transformed target space points, and the source space points: 
\[k(x,y)=\exp \Bigg( -\frac{|T(x)-y|^2}{2\sigma^2} \Bigg)\]
With some arbitrary $\sigma$. For any point $y\in Y$, we quantify how much it is contained in the distribution of $X$ by computing the mean similarity between $y$ and every other point in $X$: 
\[\text{KernelSum}(y) = \int_X K(x,y)dx=|A|^{-1}\sum_{x\in A\subset X} K(x,y) \]

\subsection{Missing Class Detection}
\begin{figure}
    \centering
    \includegraphics[width=0.5\linewidth]{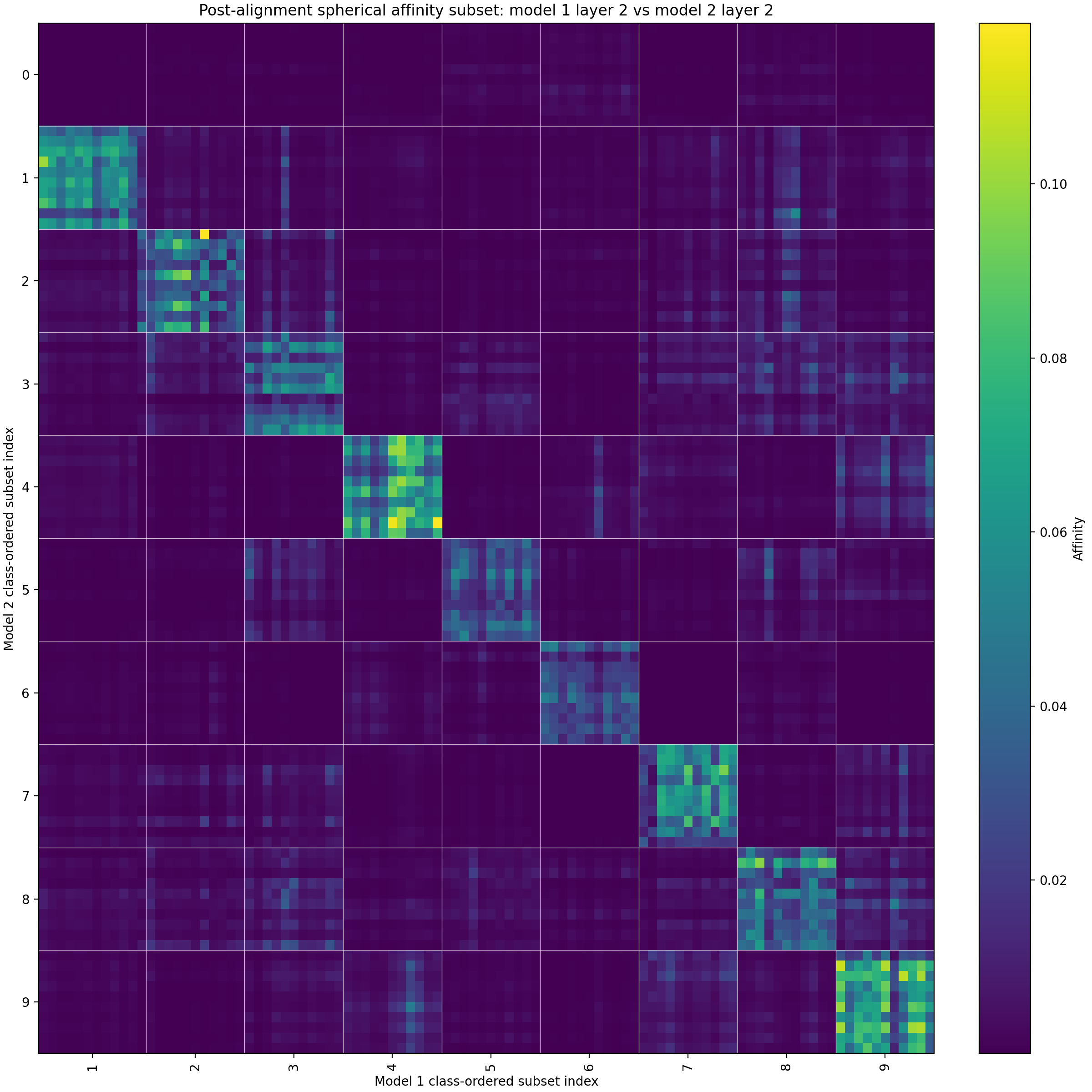}
    \caption{Subset of kernel between a model trained on MNIST excluding the ``0'' class, and a regular model trained on MNIST}
    \label{fig:excluded_class_kernel}
\end{figure}

\begin{figure}
    \centering
    \begin{subfigure}[b]{1.0\linewidth}
        \includegraphics[width=\linewidth]{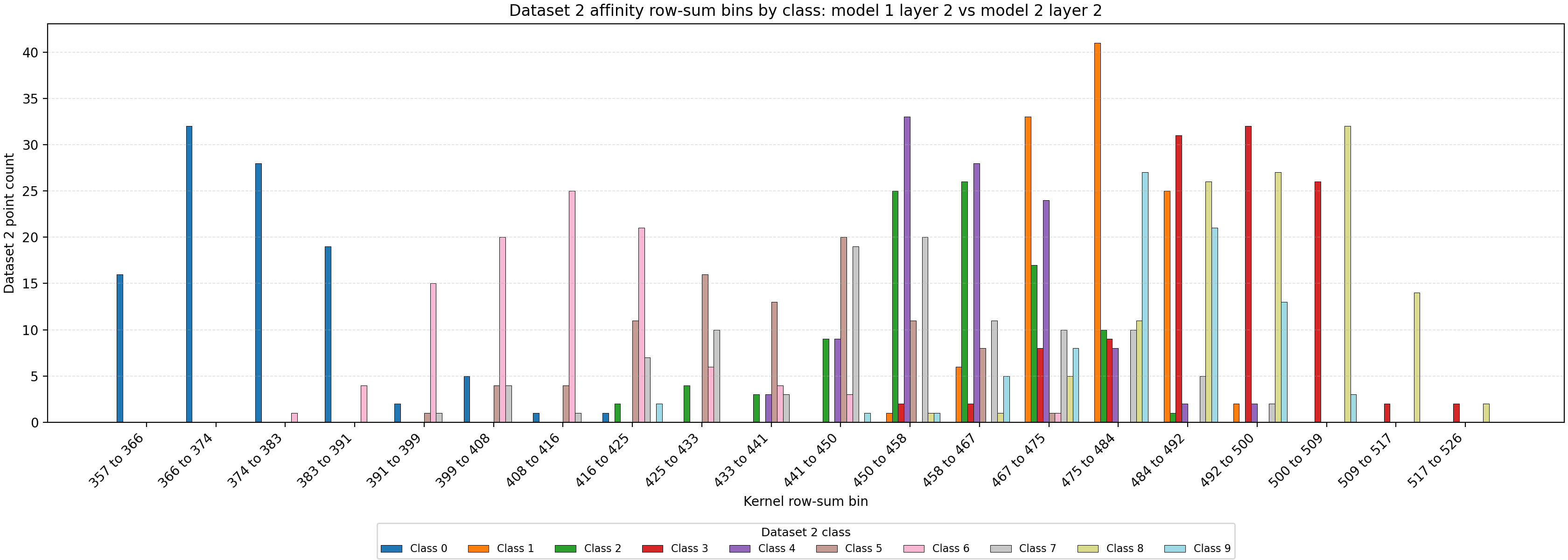}
        \caption{Ortho}
        \label{fig:excluded_class_kernel_sum_ortho}
    \end{subfigure}
    \hfill
    \begin{subfigure}[b]{1.0\linewidth}
        \includegraphics[width=\linewidth]{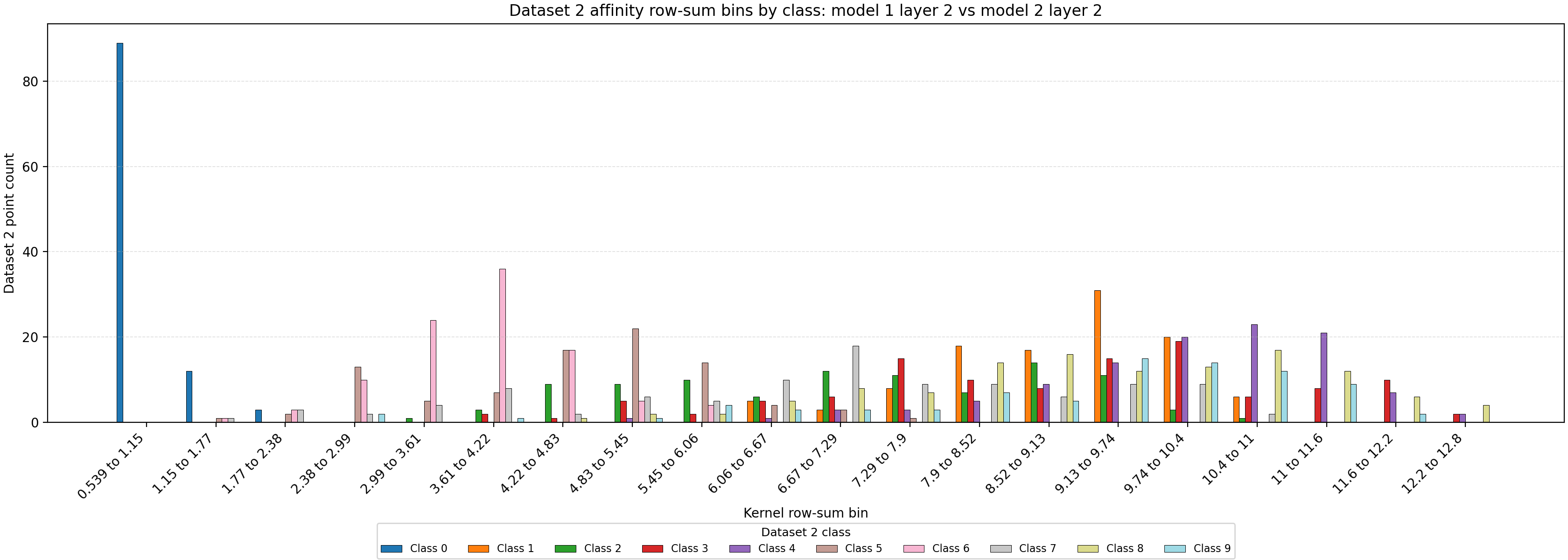}
        \caption{HGA}
        \label{fig:excluded_class_kernel_sum_hga}
    \end{subfigure}
    \caption{Kernel sum distributions for each class between a model trained on MNIST excluding the ``0'' class, and a regular model trained on MNIST}
    \label{fig:excluded_class_kernel_sum_distribution}
\end{figure}

The first experiment with this involves detecting points that have trained representations in one neural network, but not the other.
We trained a model to perform image classification on the MNIST dataset, however the ``0'' class was excluded from the dataset.
Then we tested various alignment methods between the latent space of the missing class model, and a regular model trained on MNIST with all of the anchors included.
By observing Figure \ref{fig:excluded_class_kernel}, we can see that samples from the missing class have very low kernel values.  
Visibly, samples in the target latent space which correspond to the missing class do not have similarity to any points in the source space. 
Furthermore, by viewing \ref{fig:excluded_class_kernel_sum_distribution}, we can see that for multiple alignment methods, the kernel sum values for the excluded class are much lower than the regular classes.
Using this kernel sum metric, we can effectively locate a region of the latent space geometry which is present in the target latent space, but not the source latent space. 

\subsection{Image Embedding Model Alignment}
An interesting qualitative result is the comparison of the latent spaces for image embedding models. 
We find that we can qualitatively locate differences in the embedding spaces of different image encoder models.
In this setting, we use an image encoder which was trained exclusively on image data (DINO-V2 \cite{oquab2023dinov2}) as our source space, and an image encoder which was trained with image and text data (CLIP-image \cite{radford2021learningtransferablevisualmodels}) as our target latent space. 
We generate embeddings for images from the THINGS \cite{Hebart2019THINGS} dataset, and compute the kernel sum score for each image in the target (CLIP-image) space.
By observing Table \ref{tab:lowest_kernel_concepts}, we see that concepts with the lowest kernel sums tend to be the most related to textual information. 
Intuitively, this makes sense, because the latent space for the multimodal model (CLIP-image) can be expected to have more unique representations for textual data, compared to the model trained entirely on image embeddings. 

\begin{table}[htbp]
\centering
\caption{Concepts from the THINGS dataset with the lowest average kernel sums. We show results using three different alignment methods (ranked ascending).}
\label{tab:lowest_kernel_concepts}
\begin{tabular}{clll}
\toprule
Rank & Orthogonal Procrustes & Latent Functional Maps & Spherical Alignment \\
\midrule
1  & banner          & pennant         & banner          \\
2  & pennant         & banner          & pennant         \\
3  & minivan         & tag             & scoreboard      \\
4  & scoreboard      & jersey          & tag             \\
5  & tag             & game            & jersey          \\
6  & woman           & visor           & board game      \\
7  & jersey          & scoreboard      & minivan         \\
8  & board game      & tattoo          & game            \\
9  & tugboat         & board game      & visor           \\
10 & football helmet & patch           & woman           \\
11 & game            & cleat           & football helmet \\
12 & visor           & face mask       & tattoo          \\
13 & patch           & sticker         & cleat           \\
14 & cleat           & juicer          & cassette        \\
15 & tattoo          & woman           & patch           \\
16 & seaplane        & lip gloss       & badge           \\
17 & bobsled         & football helmet & diskette        \\
18 & van             & juicer          & tugboat         \\
19 & gravestone      & minivan         & t-shirt         \\
20 & diskette        & spark plug      & van             \\
\bottomrule
\end{tabular}
\end{table}

\newpage

\end{document}